\documentclass[10pt,letterpaper]{article}

\usepackage{cogsci}

 \cogscifinalcopy 

\usepackage[
  style=apa,
  natbib=true,
  annotation=false,
]{biblatex}
\usepackage{float} 

\title{The Dual Role of Abstracting over the Irrelevant in Symbolic Explanations: Cognitive Effort vs. Understanding}

\author[1]{\mbox{Zeynep G. Saribatur (zeynep.saribatur@tuwien.ac.at)}}
\author[2]{\mbox{Johannes Langer}}
\author[2]{\mbox{Ute Schmid}}
\affil[1]{Institute of Logic and Computation, TU Wien}
\affil[2]{Cognitive Systems, University of Bamberg}

\usepackage{amsthm}
\usepackage{caption}
\usepackage{subcaption}
\usepackage{todonotes}
\usepackage{graphicx}

\def\ba{\begin{array}}
\def\ea{\end{array}}
\def\be{\begin{enumerate}}
\def\ee{\end{enumerate}}
\def\bi{\begin{itemize}}
\def\ei{\end{itemize}}
\def\beq{\begin{equation}}
\def\eeq#1{\label{#1}\end{equation}}
\def\beeq{\begin{equation*}}
\def\eeeq{\end{equation*}}

\def\mi#1{\mathit{#1\/}}
\def\AS{\mathit{AS}}

\newcommand{\Lits}{\mathcal{U}}
\newcommand{\nop}[1]{}

\newtheorem{defn}{Definition}
\newtheoremstyle{example}
  {3pt}
  {3pt}
  {\normalfont}
  {}
  {\bfseries}
  {.}
  {.5em}
  {}

\theoremstyle{example}
\newtheorem{exmp}{Example}

\begin{document}

\maketitle

\begin{abstract}
Explanations are central to human cognition, yet AI systems often produce outputs that are difficult to understand. While symbolic AI offers a transparent foundation for interpretability, raw logical traces often impose a high extraneous cognitive load. We investigate how formal abstractions, specifically removal and clustering, impact human reasoning performance and cognitive effort. Utilizing Answer Set Programming (ASP) as a formal framework, we define a notion of irrelevant details to be abstracted over to obtain simplified explanations. Our cognitive experiments, in which participants classified stimuli across domains with explanations derived from an answer set program, show that clustering details significantly improve participants’ understanding, while removal of details significantly reduce cognitive effort, supporting the hypothesis that abstraction enhances human-centered symbolic explanations. 

\end{abstract}

\section{Introduction}
Symbolic AI refers to the methods in AI that are based on explicitly describing the knowledge of the world and the problem through logical or related formal languages, and finding possible solutions through search and reasoning. Unlike subsymbolic systems (largely neural networks), symbolic and rule-based representations are inherently more transparent, offering a strong foundation for explainable AI (XAI).
Despite the dominance of deep learning, the need for symbolic methods has never been more critical. Purely statistical models, such as Large Language Models (LLMs), frequently suffer from black-box opacity, making their internal decision-making processes impossible to audit. In high-stakes environments, such as healthcare, autonomous aviation, and law, this lack of transparency is a prohibitive barrier to trust. Furthermore, statistical systems often fail at multi-step logical reasoning and are prone to hallucinations, where they generate plausible-sounding but factually incorrect information. Symbolic AI provides the necessary guidance by encoding explicit rules, ensuring that system behavior remains within defined ethical, safety, and logical boundaries. 

This synergy is central to the emerging direction of Neuro-symbolic AI which aims to bridge Kahneman’s System 1 (fast, intuitive thinking) and System 2 (slow, logical deliberation) \citep{kahneman2011thinking}. Neuro-symbolic systems use neural layers for perception—such as interpreting visual data or natural language—and symbolic layers for high-level reasoning and decision-making \citep{mao2019neuro}, which is shown to have very successful applications while also obtaining explanations~\citep{DBLP:series/faia/342}.

One challenge in adopting a symbolic view on understandability and explainability is that providing humans with rules and axioms may not be enough to achieve a clear understanding of how a system reached a decision. Initial steps towards providing explanations have been by showing the trace of rules that were applied in computing a decision \citep{clancey1983epistemology}. However, if the decision-making rules
are too large and contain many distracting details, then it is challenging for humans to understand the decision process. In the cognitive science of explanation, it is well-established that humans prefer simple, contrastive explanations over exhaustive traces \citep{miller2018explanation, lombrozo2007simplicity}. 
Just as image classification explanations use saliency maps to highlight relevant pixels while treating the rest as irrelevant \citep{ribeiro2016should}, symbolic representations must distinguish between essential logical pivots and distracting details adhering with Grice's Maxim of Quantity \citep{grice1975logic}.

The role of abstraction is thus central to making complex systems interpretable, with established theories spanning both symbolic and subsymbolic domains. In symbolic reasoning, abstraction often involves forgetting or projecting away non-essential details to distill the essence of an explanation  \citep{siebers2019please,DBLP:conf/ijcai/ChakrabortiSK20}, as well as simplifying solution spaces to foster better human-AI alignment \citep{ho2022people,poesia2022left}. These structural refinements are frequently achieved through predicate invention to streamline rule-based representations \citep{schmid2016does,muggleton2018ultra,furnkranz2020cognitive} or domain clustering to group related logical entities into higher-level concepts \citep{zgs19xai}. By reducing the granularity of the data, these methods aim to get a symbolic output that is not just technically sound, but also cognitively manageable for a human observer. Beyond purely symbolic systems, abstraction serves as a bridge for interpreting neural architectures, utilizing causal abstractions \citep{geiger2021causal}  or distilling opaque weights into readable decision trees \citep{confalonieri2021using}.

In this work, we focus on Answer Set Programming (ASP) \citep{aspglance11}, a prominent logic-based formalism in symbolic AI known for its declarative expressivity and efficient solvers. ASP is a highly popular language widely used not only in AI but in Computer Science to solve a variety of problems such as combinatorial optimisation, logical reasoning, planning, bioinformatics, and data integration to name a few
\citep{schaub2018special}, and has recently shown great potential in neuro-symbolic reasoning \citep{eiter2022neuro,DBLP:journals/tplp/BarbaraGLMQRR23}
and as a reasoning layer for LLMs \citep{kareem2024using,DBLP:journals/corr/abs-2302-03780}. 
While obtaining explanations for solutions (i.e., answer sets) of an answer set program  is a well-studied topic, achieving concise explanations that aid human understanding remains a challenge \citep{DBLP:journals/tplp/FandinnoS19}. Recent theoretical work in ASP has explored \emph{removal} \citep{zgsswkr23}  and \emph{clustering}  \citep{DBLP:conf/kr/Saribatur00024} of irrelevant details, via preserving the correspondence of answer sets for any potential extensions of the program, but these formalisms are too restricted to bring to applications and, more importantly, lack empirical validation regarding their impact on human cognition.

Our contributions are as follows:
We propose a notion of abstraction in ASP based on irrelevancy within a problem space, by relaxing the previous restricted notions.
We then use these abstractions to obtain explanations on the decision-making. We empirically evaluate how removal and clustering of irrelevant details help in human understanding and cognitive effort. Our results demonstrate a double benefit: clustering details significantly improves performance (accuracy), while removal of irrelevant details significantly reduces cognitive effort (answer time).

The rest of the paper is organized as follows. We begin with a high-level background on ASP and explanations in ASP. We then present the notion of abstracting over details that are irrelevant w.r.t. a set of potential problem instances and illustrate its use in obtaining simplified explanations. Then we describe our empirical study, and present the results. We then conclude with a discussion on future work.

\section{Background}

\subsubsection{Answer Set Programming (ASP)}

ASP is a popular declarative modeling and problem solving framework in artificial intelligence, and generally in computer
science, with roots in non-monotonic
logic~\citep{aspglance11}.

The problem specifications are described by a set
of rules using propositional atoms, meaning that
an atom $a$ can either be true or false, and the solutions to the problem instances get represented by the models (i.e., answer sets) of
the program. The roots of ASP come from formalisms that aimed at representing and reasoning about commonsense knowledge and beliefs of agents, by relying on the ability to represent non-monotonic reasoning, which is the ability to withdraw previous conclusions about the world when receiving new information. Readers interested in the use of the ASP paradigm in modeling human reasoning principles can see~\citep{DBLP:conf/cogsci/DietzFH22}. Here, we focus on the use of ASP for logical reasoning in AI with real-world applications~\citep{DBLP:journals/aim/ErdemGL16}. 

Let us present a toy example that shows the expressiveness of ASP in modeling problems, through the ``guess and check" approach. The idea is to use choice rules to guess for solutions to the problem, which are then pruned by the constraints of the problem that a correct solution should not satisfy.

\begin{exmp}\label{ex:block}
Consider the blocksworld planning problem, where from a given initial state, the aim is to find a sequence of actions, i.e., plan, that arranges the blocks so that the desired goal state is reached. 

We begin by guessing potential actions for each time step. The choice rule
$$1 \{\mi{occurs}(A,T): \mi{action}(A) \} 1 \leftarrow \mi{step}(T)$$
guesses for each step $T$ the occurrence of exactly one action $A$. 
The effect of taking an action is described with the rule
$$\mi{holds}( \mi{on}(B,L),T) \leftarrow \mi{occurs}(\mi{move}(B,L),T)$$ 
which states that if moving block $B$ to location $L$ occurs at time $T$, 
it holds that block $B$ is on location $L$. 

In order to reason over a sequence of time steps, we specify what properties carry over time.\footnote{An action usually changes a few things but leaves most of the world state untouched. Representing this efficiently without explicitly listing everything that does not change is called the Frame Problem \citep{mccarthyhayes69}.} The following rule states that if $F$ holds at time step $T$ and in the next time step $T{+}1$ it is not the case that $F$ was made false, i.e., we do not have evidence of a change, then we can infer that $F$ holds at $T{+}1$.
$$\mi{holds}(F,T{+}1) \leftarrow \mi{holds}(F,T) , \mi{not}\ \neg\mi{holds}(F,T{+}1) , \mi{step}(T).$$ 
The executability conditions for actions 
are defined using constraints, which are rules of the form
$$\bot \leftarrow \mi{occurs}(\mi{move}(B,C),T), \mi{unclear}(B,T)$$ 
that forbids the movement of block $B$ at time step $T$ if it is not clear.\footnote{In ASP, constraints, which are rules with $\bot$ (falsum) in the head denoting a contradiction, are important as they eliminate non-valid solutions. If the guessing rule generates a plan where block $a$ is moved while block $b$ is on top of it, this constraint removes that entire sequence from the list of possible solutions.} Here $\mi{unclear}(B,T)$ is an auxiliary predicate defined with the rule $\mi{unclear}(C,T) \leftarrow \mi{holds}(\mi{on}(B),C,T)$ stating that a block is unclear if it has another block on top. In order to enforce that the goal condition is reached in the computed plan, constraints are used, e.g.,
$$\bot \leftarrow \mi{goal}(\mi{on}(B,L)), \mi{not}\ \mi{holds}(on(B,L),T)), \mi{maxTime}(T).$$
\end{exmp}

\begin{figure}
    \centering
\begin{tikzpicture}[
    scale=0.6, 
    block/.style={draw, fill=gray!30, minimum width=0.8cm, minimum height=0.4cm, font=\bfseries\small},
    table/.style={draw, fill=brown!50, minimum width=2.5cm, minimum height=0.1cm},
    state/.style={align=center, font=\normalsize\bfseries},
    caption/.style={align=center, text width=3.5cm},
]

\node[state] at (0, 3.0) {Initial State ($T=0$)};
\node[table] (table1) at (0, 0) {};

\node[block] (C1) at (-0.8, 0.4) {C};
\node[block] (B1) at (-0.8, 1.1) {A};
\node[block] (A1) at (0.8, 0.4) {B};

\node[font=\tiny, align=left] at (1.2, 2.5) {$\mi{holds}(\mi{on}(a, c), 0)$};
\node[font=\tiny, align=left] at (1.45, 2.2) {$\mi{holds}(\mi{on}(c, \mi{table}), 0)$};
\node[font=\tiny, align=left] at (1.45, 1.9) {$\mi{holds}(\mi{on}(b, \mi{table}), 0)$};


\node[draw, rectangle, fill=white, inner sep=1pt, font=\tiny, align=left] at (4, 0.5) {
    \textbf{Plan Steps:}\\
    1. move(a,table)\\
    2. move(b, c)\\
    3. move(a, b)
};

\node[state] at (8, 3.0) {Goal State ($T=3$)};
\node[table] (table2) at (8, 0) {};

\node[block] (B2) at (7.2, 0.4) {C};
\node[block] (C2) at (7.2, 1.1) {B};
\node[block] (A2) at (7.2, 1.8) {A};

\node[font=\tiny, align=left] at (9.3, 2.5) {$\mi{goal}(\mi{on}(a, b))$};
\node[font=\tiny, align=left] at (9.3, 2.2) {$\mi{goal}(\mi{on}(b, c))$};
\node[font=\tiny, align=left] at (9.55, 1.9) {$\mi{goal}(\mi{on}(c, \mi{table}))$};

\end{tikzpicture}
\caption{Blocksworld example}
\label{fig:bl}
\end{figure}

Once we have an ASP description of the problem, we can use an ASP system (e.g., Clingo~\citep{DBLP:journals/tplp/GebserKKS19}) to solve a given problem instance described through a set of facts. In case of a planning problem, these facts describe the initial state and the desired goal state.
The answer sets 
of the program then contain the solutions to the problem.

\begin{exmp}[Ex.~\ref{ex:block} cont.]
For initial and goal states shown in Figure~\ref{fig:bl}, the answer set contains
$\mi{occurs}(\mi{move}(a,\mi{table}), 1)$, $\mi{occurs}(\mi{move}(b, c), 2)$,  $\mi{occurs}(\mi{move}(a, b), 3)$ which describes the solution to the planning problem.
\end{exmp}

\paragraph{Explanations in ASP} The main focus on explanations in ASP is explaining the reasoning behind an obtained answer set. In particular, given an answer set $I$ of a program $P$, a user might ask questions about the presence or absence of certain atoms in $I$. Interested readers are refered to \citet{DBLP:journals/tplp/FandinnoS19} for an overview of the explanation approaches. The main idea is to provide a justification of the reached solution via tracing the relevant rules/atoms and showing them through a graph or a tree structure. There are several tools that can be used for this purpose, including a recent one \texttt{xclingo} \citep{DBLP:journals/tplp/CabalarM24}. For example, for Example~\ref{ex:block} depending on how the problem is encoded \citep{cabalarACLAI}, the explanation for the atom $h(\mi{on}(a),b,3)$ to hold true in the answer set is a trace with the actions $o(\mi{move}(a,b),3)$  $o(\mi{move}(a,\mi{table}),1)$ and the changes of locations of block $a$.

However, as the complexity of the domain rises, these explanations suffer from information density. Consider a scenario where only colored blocks can be moved. If all blocks in every possible initial state are colored, the \emph{colored} attribute would be part of the logical trace while providing no discriminative detail. To a human observer, these redundant details are distracting and increases cognitive load. This motivates the need for formal notions of irrelevancy to be removed or clustered over, which we define in the next section.

\section{Abstracting the Irrelevant in Problems}

To systematically reduce the complexity of explanations, we must first define which parts of a logical program are "safe" to simplify. We focus on abstracting over details that are irrelevant for decision-making within specific problem contexts. For this, we relax previous notions of abstractions \citep{zgsswkr23,DBLP:conf/kr/Saribatur00024}, by focusing on the consistency of answer sets under a set of potential scenarios.


 \begin{defn}[$\chi$-irrelevance]\label{def:irrelevance}
Given a program $P$ over $\Lits$ and a set $\chi = \{I_1,\dots,I_n\}$ with $I_i \subseteq \Lits$, the sets $A_1,\dots,A_n \subseteq \Lits$ of atoms are \emph{$\chi$-irrelevant} if for a (surjective) mapping $m: \Lits \rightarrow \Lits' \cup \top$ with $|m(A_i)|=1, 1\leq i \leq n$, for any $F \in \chi$
\beq 
m(\AS(P \cup F)) = \AS(m(P) \cup m(F)).
\eeq{eq:ir-cl} 
Here, $m(P)$ is the \emph{($\chi$-)abstracted program}.
 \end{defn}
 
The aim is to define atoms that are irrelevant in $P$ w.r.t. the provided problem instances, via the existence of an (abstracted) program that can be obtained with applying a mapping to $P$. Here, the mapping $m$ performs two primary cognitive operations: (i) Removal, which is mapping atoms to $\top$ (logical truth), represents \emph{forgetting} a detail that lacks discriminative power, and (ii) Clustering: Mapping multiple distinct atoms to a single representative atom, represents \emph{chunking} and reducing the granularity of the domain.

To illustrate how these notions relate to obtaining explanations, consider a toy classification task.

\begin{exmp}\label{ex:toy}
Let us consider the program $P$
\begin{align*}
    \mi{needsWater} \leftarrow \mi{habitatWater}\\
    \mi{needsWater} \leftarrow \mi{habitatMud}\\
\mi{scent} \leftarrow \mi{spiky}, \mi{needsWater}, \mi{headLargerLeaf}
\end{align*}
which states if the flower has a larger head than its leaves, if if lives in a habitat which shows it needs water, and it if has spiky leaves then we can classify it as having a scent.

The set  $\chi=\{\{\mi{spiky},\mi{habitatWater},\mi{headLargerLeaf}\},$ $\{\mi{spiky},\mi{habitatMud}, \mi{headSmallerLeaf}\},\{\mi{spiky},\mi{habitatSand},$ $\mi{headLargerLeaf}\}\}$ 
presents three flowers. All these flowers have spiky leaves, but they vary in their habitat, and head size vs. leaf size.

Observation on \emph{removal}: Since $\mi{spiky}$ are true for all the flowers, it is not decisive in classification, 
    and can be removed ($m(\mi{spiky}) = \top$). Observation on \emph{clustering}: Since both, $\mi{habitatWater}$ and $\mi{habitatMud}$ (but not $\mi{habitatSand}$), reach to the same outcome $\mi{needsWater}$, 
    they can be clustered into a single abstract concept $m(\mi{habitatWater})=m(\mi{habitatMud})=\mi{habitatWaterOrMud}$  
    Applying the mapping $m$
    yields the program $m(P)$ below.
	\begin{align*}
    \mi{needsWater} \leftarrow \mi{habitatWaterOrMud}\\
\mi{scent} \leftarrow \mi{needsWater}, \mi{headLargerLeaf}
	\end{align*}
    It can be observed that $m(P)$ satisfies \eqref{eq:ir-cl} for any $F\in \chi$.
\end{exmp}

\begin{figure}[t]
    \centering
   \begin{subfigure}[b]{0.20\textwidth}
    \centering
   \footnotesize
\begin{verbatim}
|__scent                  
| |__spiky            
| |__headLargerLeaf            
| |__needsWater     
| | |__habitatWater
\end{verbatim}
\caption{Default}
\label{fig:def}
\end{subfigure}
\begin{subfigure}[b]{0.25\textwidth}
    \centering
\footnotesize
\begin{verbatim}
|__scent                  
| |__headLargerLeaf            
| |__needsWater     
| | |__habitatWaterOrMud
\end{verbatim}
\caption{Abstracted (cluster\&removal)}
\label{fig:abs}
\end{subfigure}

    \caption{Explanations obtained by \texttt{xclingo}}
    \label{fig:enter-label}
\end{figure}

We are interested in making use of this notion to generate more concise justifications. The idea is to obtain justifications from the abstracted program, i.e., $m(P)$. Using the  \texttt{xclingo} system \citep{DBLP:journals/tplp/CabalarM24}, we can visualize the difference between a default justification and an abstracted one.

\begin{exmp}[Ex.~\ref{ex:toy} ctd]
Let us look at the scenario $\chi_1=\{\mi{spiky},\mi{habitatWater},\mi{headLargerLeaf}\} \in \chi$. As seen in Figures \ref{fig:def} and \ref{fig:abs}, the default explanation for the atom $\mi{scent}$ to occur in the answer set of $P\cup \chi_1$ includes the $\mi{spiky}$ leaf and the specific habitat. The abstracted explanation for $\mi{scent}$ appearing in the answer set of $m(P) \cup m(\chi_1)$, however, prunes the redundant detail and generalizes the habitat.
\end{exmp}

Now, with our formal notions at hand, we can ask the central question of our empirical study: Do these theoretically simpler traces actually result in measurable improvements in human understanding and reductions in cognitive effort?

\section{Empirical Study}

We conduct an empirical study to observe whether abstract explanations help in human understandability, cognitive effort and confidence. The idea is to present the participants with explanations of a reasoning task, and then evaluate their performance for determining the outcome when encountered with new instances. For measuring understanding and effort, we look at accuracy and answer times. Subjective confidence is assessed via participant self-reports collected after each classification task.

We formulate our research questions as follows:

\noindent \textbf{Q1.}\enspace Do explanations abstracted over irrelevant details 
improve participant accuracy when performing transfer tasks on new instances?

As with abstraction by removal and by clustering, we obtain simplified explanations with only the relevant details that reaches the outcome, we hypothesize that these explanations allow participants to better understand the underlying logic of the reasoning trace, leading to fewer errors when applying that logic to novel scenarios. 

\noindent \textbf{Q2.}\enspace Do explanations with irrelevant details removed 
reduce the answer time when deciding on the new instances?


The aim of the removal abstraction is to keep the decisive details. We expect that participants will spend less time parsing redundant information, leading to more efficient decision-making and faster answer times.

\noindent \textbf{Q3.}\enspace Do explanations abstracted over irrelevant details 
increase human confidence in their decision making?

If an explanation is concise and highlights only the relevant path to an outcome, we expect the participants feel more confident in their understanding, correlating with their objective improvements in performance (Q1 and Q2).

\subsection{Method}
\subsubsection{Task}
\begin{figure}[t]
    \centering
    \includegraphics[width=1\linewidth]{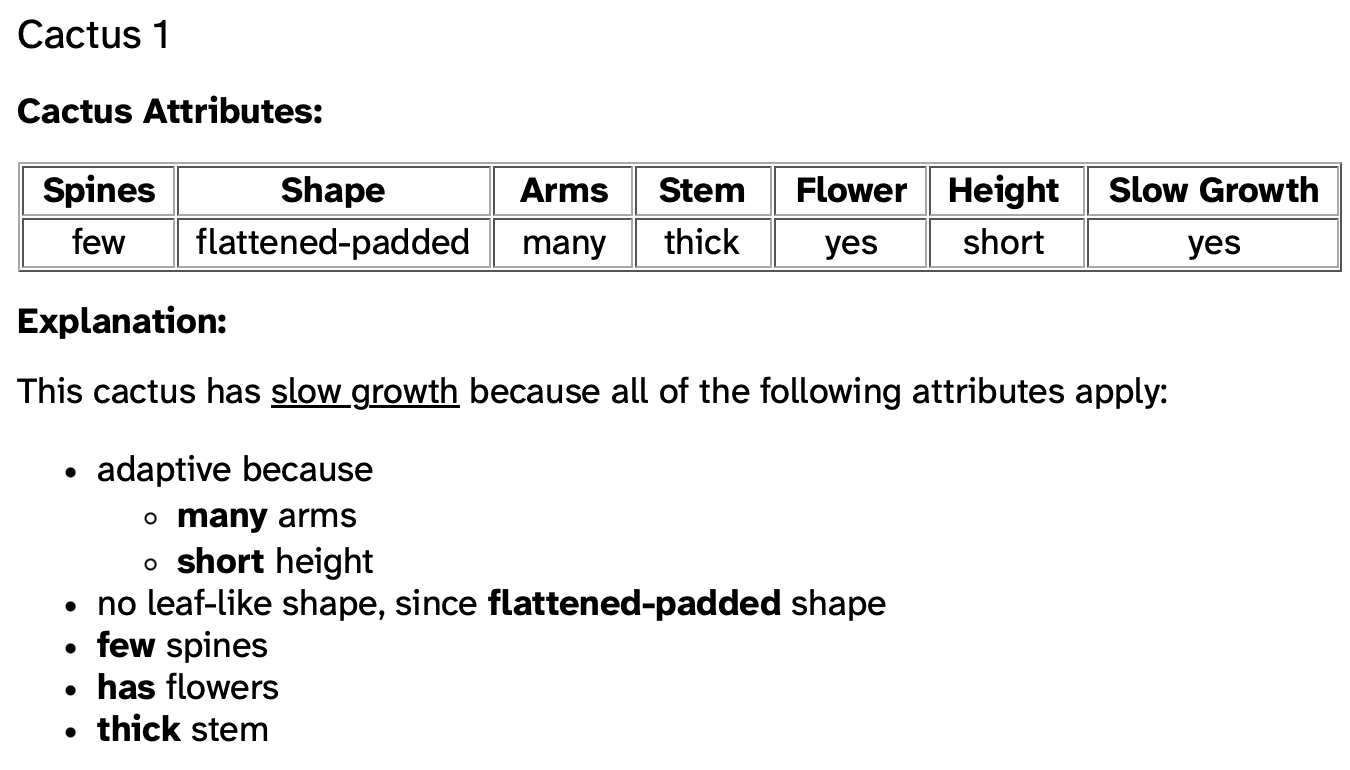}
    \caption{Example of a learning phase instance as presented in the study. The table 
    lists all attributes and values. 
    The provided explanation is the default without any abstraction. 
     }
    \label{fig:stimulus}
\end{figure}

The experiment utilized concept-learning tasks in which participants performed a binary classification task after being presented with explanations (e.g., Figure~\ref{fig:stimulus}). 
Explanations in the context of symbolic AI approaches rely on human-understandable concepts and relations between them \citep{poeta2023concept}. As such, they are closely related to concept learning \citep{lake2015human}. In the context of rule-based classification, concept-based explanations support humans in understanding the relevant aspects of instances to make them belong to a specific class or category. This has been shown, for instance, in the context of inductive logic programming \citep{muggleton2018ultra,rabold2022generating}.

\emph{Domains.}
For the classification task, 
we designed three different domains, each of which being of different biological specimen: flowers, mushrooms, and cacti. Each domain has six domain-specific decision attributes and one binary target label. We used domain-specific neutral target terms, such as flower having a \emph{scent} in contrast to it being \emph{poisonous}, with no implied consequence to the decision. We decided on having tasks from similar but different domains in order to prevent proactive interference~\citep{underwood1957interference}, where learning one rule hinders the ability to learn the next.  
 We designed 8 instances for each domain: 2 positive instances to be used for the learning phase, and 3 positive/3 negative instances for the test phase. 

\emph{Classification rules.} Each classification task is based on an answer set program representing the conditions needed to reach the target class. As a representative we show below the program for the cactus domain which determines when a cactus has \emph{slow-growth}.\footnote{Here $-\mi{shape}(X,\mi{leaflike})$ is true whenever $\mi{shape}(X,\mi{cylindrical})$ or $\mi{shape}(X,\mi{padded})$ is true. We use the notation for \emph{strong negation} ($-$), since \texttt{xclingo} does not provide explanations of default negation ($\mi{not}$), due to its underlying theory.}
\begin{align*}
\mi{adaptive}(X)  \leftarrow& \mi{arms}(X,C), \mi{height}(X,\mi{short}).\\
\mi{slowGrowth}(X)  \leftarrow& \mi{adaptive}(X), \mi{spines}(X,C), \mi{flower}(X,\mi{yes}),\\
&-\mi{shape}(X,\mi{leaflike}), \mi{stem}(X,\mi{thick}).
\end{align*}
Figure~\ref{fig:stimulus} shows an example explanation for a cactus to be classified as having \emph{slow-growth}.
The program for the flower domain extends that from Example~\ref{ex:toy}, and the mushroom domain is represented similarly to define the conditions for a mushroom to be \emph{tolerant}. We calculated for each answer set program of the three domains the $\chi$-irrelevant atoms where $\chi$ is the set of all positive/negative
instances of the respective domain, for removal and for clustering. We use the abstracted answer set programs to obtain the abstracted explanations. 


\begin{figure}[t]
    \centering
    \includegraphics[width=0.75\linewidth,trim=3em 2em 3em 0]{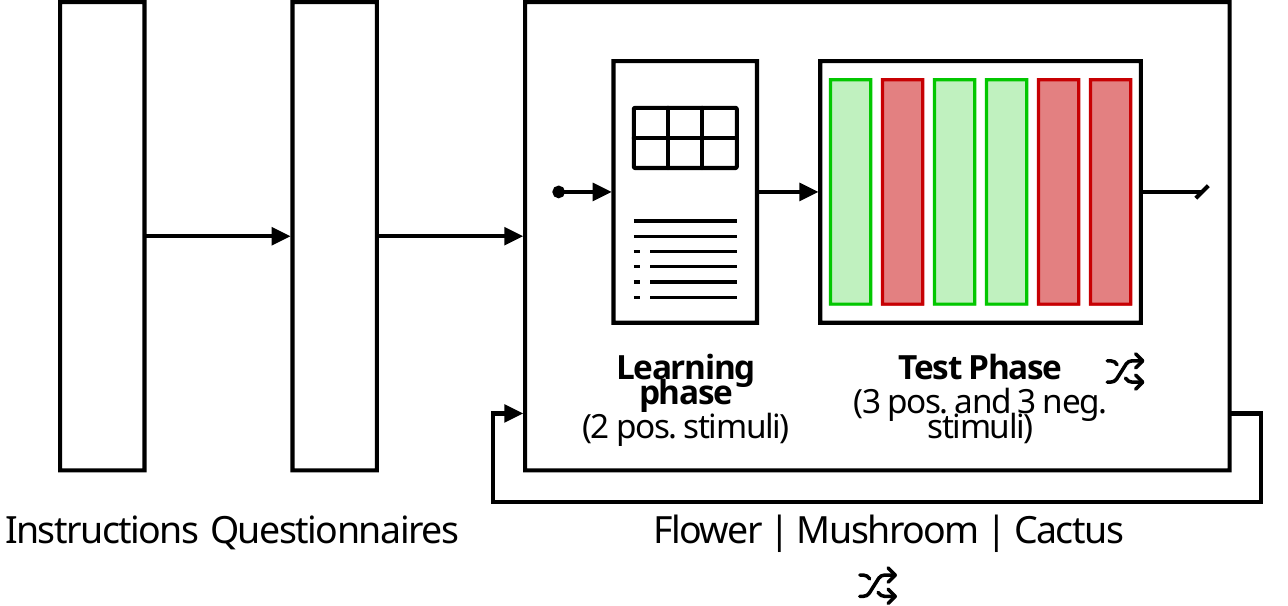}
    \caption{Study Procedure Overview. The order of the three domains and the test phase, i.e. the order of the positive (illustrative in green) and negative (resp. in red) stimuli were random to balance out possible position effects.}
    \label{fig:study_procedure}
\end{figure}

\paragraph{Study Design and Procedure}
The empirical online study is based on a complete between-subject design, where participants were randomly assigned to one of the four groups: \texttt{default}, \texttt{cluster}, \texttt{removal}, and \texttt{cluster and removal}, and the data between these groups were compared to calculate the effect on our dependent variables.\footnote{Informed consent about data protection, anonymity and the right to leave the study at any time was given. At the end of the study, participants had the opportunity to leave further comments and notes.} The 
experiment 
consists of a learning and a test phase (see Figure~\ref{fig:study_procedure}).

 \emph{Learning Phase.} At the beginning of each domain, participants receive 2 positive examples together with an explanation, a textual form of the output from \texttt{xclingo}, on why the instance was classified as such (see Figure~\ref{fig:stimulus}).
%
%
These explanations differ depending on the assigned group of the experiment. The \texttt{default} group receives a full justification trace including all technical attributes leading to the classification 
(see Figure~\ref{fig:stimulus}). 
The \texttt{cluster} and \texttt{removal} groups receive abstracted explanations based on clustering (e.g. 'water or mud') or removal of $\chi$-irrelevant atoms, respectively. 
The \texttt{cluster\_removal} group sees abstracted explanations resulting from the application of both strategies (e.g., Figure~\ref{fig:abs}).

\emph{Test Phase.} After the learning phase, unseen instances (3 positive, 3 negative) are presented without explanations. Participants need to decide \emph{stimulus valence} that is whether the instance is classified as positive or not and give a rating about their confidence with a slider bar (range 0-100), or choose the option ``I don't know''. 
The attribute values were evenly distributed across all instances to avoid accumulations of a single attribute value.

\emph{Subjective Assessments.} After the last domain, participants gave ratings on the perceived usefulness of the explanations (5 items, based on a subset of \cite{bahel24}), their memorization efforts (1 item) and their familiarity with the domain terminologies (1 item). All these items were rated on a Likert-type scale ranging from 1 ('strongly disagree') to 7 ('strongly agree'). Prior knowledge in computer science, logic, mathematics and programming as well as the three domains (7 items) is rated on a scale from 0 ('none') to 4 ('expert').

%

\paragraph{Participants}
We recruited participants via the online platform \textit{Prolific}. In order to ensure that performance metrics reflected genuine reasoning rather than fatigue or random guessing, we implemented comprehension checks and attention checks. We informed the participants that the survey can only be taken once, in order to avoid familiarity with the previously learned rules. The survey screened-out those that restarted the survey after failing a check. 
Among the 157 participants that have started to survey, 103 have completed the survey and the questionnaire. Furthermore, we had to exclude from the analysis, 3 participants due to restarting the survey, which missed the online checks during the survey, and 2 participants, due to low answering times (more than three standard deviations from the mean of the assigned group). The final sample size comprised of 99 participants (age mean = $35.9$, sd = $14.0$; $64$ female, $32$ male, $1$ other). Participants were randomly assigned to one of four conditions: Default ($n=19$), Cluster ($n=19$), Removal ($n=15$), or Cluster-Removal ($n=44$). While final group sizes were unequal due to unbalanced group sizes resulting from the real-time random assignment process, Levene’s test confirmed that the assumption of homogeneity of variance was maintained ($p > .05$) for two of the dependent variables (Accuracy, Confidence). For Answer Time, where the assumption was violated, we utilized the robust Welch ANOVA and Games-Howell post-hoc tests to ensure statistical validity.  There is no significant difference in the distribution of gender, age, or self-reported knowledge in computer science among the abstraction groups. Participants were paid for their participation.

\subsection{Results}

\begin{figure*}[ht!] 
    \centering
    \begin{subfigure}[b]{0.48\textwidth}
        \centering
        \includegraphics[width=1.1\textwidth]{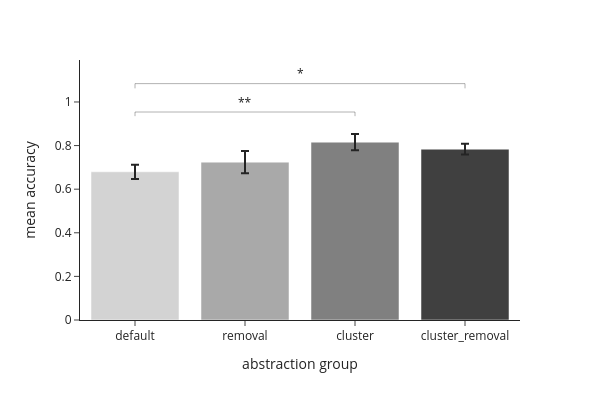}
    \end{subfigure}
    \hfill 
    \begin{subfigure}[b]{0.48\textwidth}
        \includegraphics[width=1.1\textwidth]{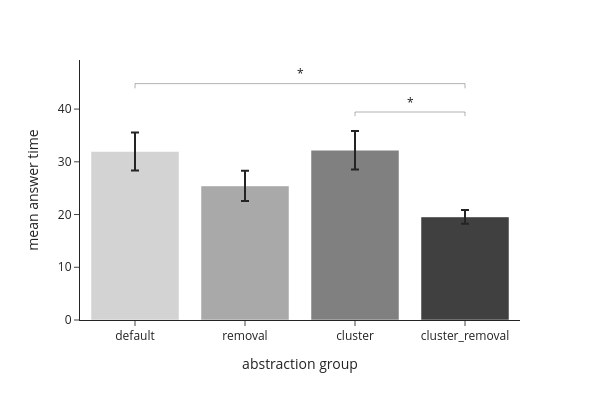}
        \centering
    \end{subfigure}
    
    \caption{Mean accuracy (left) and mean answering times (right) across the abstraction levels. Error bars represent $\pm$1 SEM. Significant effect observed for pairwise comparison between the groups are marked with $\star$ ($p < 0.05$) and  $\star\star$ ($p < 0.01$)} 
    \label{fig:combined_results}
\end{figure*}

\paragraph{Analysis}

We report our statistical tests regarding our previously formulated research questions. We tested effects for accuracy, answer time and confidence. No effects of the domain could be observed on our dependent variables. Therefore we report results aggregated over all domains (see Figure~\ref{fig:combined_results}).

%
\textbf{Q1  Accuracy:} 
A one-factor ANOVA considering the different types of abstraction shows a significant effect of abstraction on mean accuracy ($F=2.72,	p=0.049$). Posthoc pairwise comparison tests show significant effect for \texttt{cluster} vs. \texttt{default}  ($T=2.75, p=0.009$) and for \texttt{cluster\_removal} vs. \texttt{default} ($T=2.52, p=0.016$).

\textbf{Q2 Answer Time:} 
A one-factor Welch ANOVA considering the different types of abstraction shows a significant effect of abstraction on mean answer time ($F=6.41, p=0.001$). Posthoc tests show significant effect for \texttt{cluster} vs. \texttt{cluster\_removal}  ($T=3.25, p=0.017$) and for \texttt{cluster\_removal} vs. \texttt{default} ($T=-3.23, p=0.018$).

 \textbf{Q3 Confidence:}
A one-factor ANOVA shows no significant effect of abstraction over mean reported confidence ($F=1.58, p=0.19$). In the postdoc pairwise comparison tests we see marginal improvement for \texttt{default} vs. \texttt{removal} ($T=-1.71, p=0.096$) and \texttt{cluster} vs. \texttt{default} ($T=1.84, p=0.072$). Confidence was reported higher by the participants in their correct answers (mean $80 \pm 1.56$) than in their incorrect answers (mean $69 \pm 2.31$).

\subsubsection{Exploratory Findings}
The exploratory ANOVAs showed a main effect of stimulus valence ($F=9.89, p = 0.002$), where accuracy for positive stimuli ($M = 0.82 \pm 0.02$) was significantly higher than for negative stimuli ($M = 0.70 \pm 0.02$). This is inline with the findings that negative instances are more difficult  when the concept is conjunctive \citep{bourne1968learning}. No significant interaction with abstraction was observed, which suggests that these symbolic abstractions provide a consistent cognitive benefit regardless of the stimulus valence. %
Regarding subjective assessments on the usefulness of the explanations, we observe no difference between the conditions. 

\section{Discussion}

The findings confirm that abstractions help with understanding of the explanations and reduces the cognitive effort, though the type of abstraction influences the nature of the improvement. 

The results for Accuracy (Q1) show a significant effect of clustering. When explanations were abstracted into clusters, participants made significantly fewer errors. Interestingly, the way we presented the clusters, by grouping the features without providing a high-level semantic label, closely resembles the lists described by \cite{rissman2024words}. They showed that providing exemplar lists does not necessarily activate the concept (e.g., from \emph{water,mud} to \emph{wet}), though for our setting, these clusters likely provided structural organization that allowed participants to form more clear models of the classification rules, leading to fewer errors.

The lack of a significant effect of removal on accuracy suggests two distinct possibilities: (i) The original explanations may have already been below a complexity threshold for our participants, so that further pruning offered no benefit; (ii) Our formal notion of $\chi$-irrelevance for removal might be capturing \emph{specificity}~\citep{DBLP:journals/cp/BolognesiBC20}, as it removes details that are common across instances, and focuses on the distinctive details. This resembles deciding on whether a cat is Siamese by looking at its eye color, rather than existence of whiskers or paws as they are common to all cats. For the participants, this type of removal might not have been clear in the presented explanations, preventing them to generalize to new cases.

The results for Answer Time (Q2), on the other hand, show significance for removal of the irrelevant details in explanations for participants to reach decisions faster.  The finding that the \texttt{cluster\_removal} group was significantly faster than both  \texttt{default} and \texttt{cluster} groups suggests a synergetic effect. While clustering alone helps in accuracy, it is the removal of irrelevant details that acts as the primary engine for speed. Yet, clustering neither hurts response time, nor does removal of irrelevant details hurt accuracy. From a cognitive perspective, since the formal abstraction already removed the irrelevant details, they were not displayed in the explanations. This then helped in the participants to ignore the undecisive attributes when making the classification decisions.

A notable finding is the lack of a significant effect of abstraction on Confidence (Q3). Although participants in abstracted groups performed better and faster, this did not reflect in their perceived confidence. They had a reliable sense of their own reasoning accuracy, regardless of the explanation type.

\section{Conclusion}

In this work, we presented an experimental evaluation on how the formal notions of removal and clustering of irrelevant details in explanations help in human understanding and cognitive effort.  For this, we utilized ASP as a formal foundation to establish the notion of irrelevancy within a set of problem instances, by relaxing the notions of dependency-preserving abstractions. 
Our study provides empirical evidence for the double benefit of symbolic abstraction in explanations: structural organization via clustering facilitates understanding, while the removal of irrelevant details reduces cognitive effort. We observe that there is more than just obtaining \emph{simpler} explanations, as different abstraction operations serve distinct cognitive effects.

Since tasks require to be of a certain intermediate complexity for explanations to be helpful \citep{ai2021beneficial}, we plan to explore how complexity of the default explanation plays a role, for removal abstraction to become significant in accuracy, and for abstraction to significantly improve confidence. Another direction will be to investigate whether $\chi$-irrelevancy indeed captures specificity rather than abstraction.

Our findings highlight the importance of tailoring symbolic explanations to cognitive processes, bridging formal logic-based AI with theories of explanation in cognitive science. 
Our work positions ASP as a system for studying how symbolic abstraction can foster cognitively aligned AI.

\section{Acknowledgments}

This work has been supported by the Austrian Science Fund (FWF) project T-1315 (Saribatur).

\printbibliography

\end{document}